\documentclass[10pt,twocolumn,letterpaper]{article}

\usepackage[pagenumbers]{cvpr}
\usepackage[T1]{fontenc}
\usepackage{caption}

\usepackage{multirow}
\usepackage{booktabs}
\usepackage{array}

\newcommand{\parencite}[1]{\citep{#1}}

\usepackage[pagebackref,breaklinks,colorlinks]{hyperref}
\hypersetup{pdftitle={Effectiveness of Automatically Curated Dataset in Thyroid Nodules Classification Algorithms Using Deep Learning},
            pdfauthor={Jichen Yang, Jikai Zhang, Benjamin Wildman-Tobriner, Maciej A. Mazurowski}}

\begin{document}

\title{Effectiveness of Automatically Curated Dataset in Thyroid Nodules Classification Algorithms Using Deep Learning}

\author{
Jichen Yang$^{1}$\thanks{Corresponding author: \texttt{jy168@duke.edu}. Changes from version 1 are listed in Appendix~\ref{app:changes}.} \quad
Jikai Zhang$^{1}$\\
Benjamin Wildman-Tobriner$^{2}$ \quad
Maciej A.\ Mazurowski$^{1,2}$\\
{\normalsize $^{1}$Department of Electrical \& Computer Engineering, Duke University, Durham, NC, USA}\\
{\normalsize $^{2}$Department of Radiology, Duke University Medical Center, Durham, NC, USA}
}

\makeatletter
\g@addto@macro\@maketitle{%
\begin{center}
\captionsetup{type=figure}
\includegraphics[width=\textwidth,alt={Two panels. Left: horizontal stacked bars of the training images of the Manual Set (803 images; radiologist-curated, 378 patients), the MADLaP Set (5,228 images; automatic, run on all 3,981 patients) and the S1 Set (4,150 images; MADLaP Stage 1 only), split into benign and malignant images, with an arrow showing that the Manual Set annotations were used to develop MADLaP. Right: test-set AUC of every complete configuration (small markers) and, with 95 percent confidence intervals, of the three named models, the four radiologists, the average of their scores and the benchmark network.}]{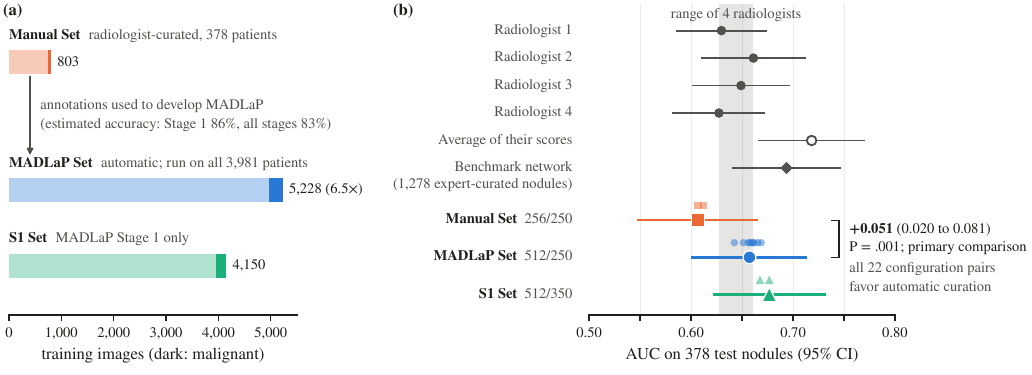}
\caption{The same expert annotations, used two ways. (a) The radiologist annotations of 378 patients were used either directly as training data (Manual Set) or to develop MADLaP, which curated the 3,981-patient development cohort (MADLaP Set; S1 Set: the output of its first stage). Evaluating MADLaP used a further 93 annotated patients \citep{zhang2023multistep}. Bars are to scale; dark segments are malignant images. (b) AUC (95\% DeLong CI) on the same 378 test nodules (147 malignant). Small markers: 10-model ensembles of every complete configuration; large markers: the primary Manual and MADLaP models (250 epochs; batch sizes 256 for Manual and 512 for MADLaP) and the S1 model with the higher cross-validation AUC (512/350); labels give batch size/epochs. Grey: four radiologists who rated the same nodules (band: their range), the average of their scores, and a benchmark network trained on 1,278 expert-curated nodules from another institution \citep{buda2019management,weng2023deep}. The radiologists' AUCs are of five-point scores; coarsened to each radiologist's rating distribution, the MADLaP model's AUC is 0.025 to 0.030 below that radiologist's, not significantly (P~=~.33 to .40; Appendix~\ref{app:sens}). Paired tests are in Tables~\ref{tab:main} and~\ref{tab:readers}.}\label{fig:teaser}
\end{center}%
}
\makeatother
\maketitle

\begin{abstract}
The diagnosis of thyroid nodule cancers commonly utilizes ultrasound images. Several studies showed that deep learning algorithms designed to classify benign and malignant thyroid nodules could match radiologists' performance. However, data availability for training deep learning models is often limited due to the significant effort required to curate such datasets. Our previous study proposed a method (MADLaP) to curate thyroid nodule datasets automatically. It was tested to have a 63\% yield rate and 83\% accuracy. However, the usefulness of the generated data for training deep learning models remains unknown. In this study, we conducted experiments to determine whether using an automatically-curated dataset improves deep learning algorithms' performance. The expert annotations of 378 patients either trained a model directly (803 images) or built MADLaP, which curated 5,228 images from 3,981 patients; the same network and training method were used throughout, and 10-model cross-validation ensembles were tested on 378 nodules from an earlier period. The model trained on the automatically-curated dataset had a significantly higher AUC than the model trained on the manually selected dataset (0.657 vs 0.607; difference 0.051, 95\% CI: 0.020, 0.081; P~=~.001), an advantage seen in all 22 configuration pairs. On the same test set, this model was within the AUC range of four radiologists (0.63--0.66) and not significantly different from a benchmark trained on 1,278 expert-curated nodules (0.69; difference $-$0.036, 95\% CI: $-$0.083, 0.011), whereas manually curated models fell significantly below it. The Stage-1 subset (86\% accuracy) did not differ significantly from the full MADLaP dataset (difference 0.015, 95\% CI: $-$0.006, 0.036). In conclusion, we showed that using an automatically-curated dataset can substantially increase the performance of deep learning algorithms.
\end{abstract}

\section{Introduction}
Ultrasound (US) remains the standard of care for imaging evaluation of thyroid nodules \citep{haugen2016american,tessler2017acr}. As is the case throughout medical imaging, artificial intelligence (AI), and particularly deep learning (DL) have been increasingly applied to thyroid US to improve diagnostic accuracy and efficiency \citep{tessler2023artificial}. Several deep learning applications have been shown to accurately characterize nodules as benign or malignant, and some have even performed better than radiologists \citep{buda2019management,li2019diagnosis,koh2020diagnosis,peng2021deep,xu2023generalizability}. DL analysis of thyroid nodules continues to be an active area of research, and a handful of commercial products have received FDA clearance for this task as well \citep{thomas2026artificial,ashton2024assessment}.

When designing deep learning tools for medical image analysis such as thyroid US, large datasets remain key for training a high-performing algorithm \citep{wang2021review,sun2017revisiting,willemink2020preparing}. However, curating and annotating large datasets of radiologic images is time-consuming and labor-intensive \citep{tajbakhsh2020embracing,willemink2020preparing}. In the case of thyroid US, data curation involves analysis of pathological and radiological data to select images of interest and assign them a pathology-based label \citep{buda2019management,zhang2023multistep}. One potential solution to this problem is to use tools and pipelines that automatically annotate images to generate larger medical datasets \citep{wang2017chestx,irvin2019chexpert,yan2018deeplesion}. Very large thyroid US datasets have been assembled at a few centers \citep{li2019diagnosis,peng2021deep}, yet deep learning for thyroid US still ``requires a significant amount of human intervention when images are selected'' \citep{koh2020diagnosis}. In our prior work,  we created an automated labeling procedure (called MADLaP) that used a combination of tools to annotate images from thyroid ultrasounds \parencite{zhang2023multistep}. The tool harnessed imaging deep learning, natural language processing, and optical character recognition to extract data from pathology reports, radiology reports, and US images. The outputs of the pipeline were labeled US images, ready for DL development.

The purpose of this study was to answer the question whether the automatically-curated datasets are indeed useful for development of accurate deep learning models. Since automatically-curated labels are imperfect, this is a question that needs to be carefully answered. Our prior work noted that the accuracy of its output may still be insufficient for tasks that require near-perfect labels \citep{zhang2023multistep}. Specifically, we aimed to evaluate how much algorithm performance might increase when trained on an automatically acquired, larger dataset compared to a model based on a smaller, manually-curated dataset. We trained several deep learning models with different compositions of datasets but the same neural network structure. As in studies of data scale and weak supervision \citep{sun2017revisiting,dunnmon2020cross}, the design is data-centric: every training set is used with the same network and training procedure, the batch size and number of epochs vary by configuration, and the training sets are compared in all 22 pairs of configurations. Finally, we explored how best to use such data: whether a smaller but more accurate subset of the MADLaP output is preferable, and whether the training batch size changes the result.

Our contributions are as follows:
\begin{itemize}
\item \textbf{A controlled comparison of two uses of the same expert annotations.} With the network, the training procedure (apart from batch size and number of epochs) and the test set held fixed, a model trained on the automatically curated MADLaP Set outperformed one trained on the expert-curated Manual Set from which MADLaP was built (AUC 0.657 vs 0.607; P~=~.001), in all 22 pairs of training configurations and with the models of all complete configurations pooled. To our knowledge, this is the first such comparison for thyroid US; because the advantage held across the batch sizes (128 to 1,024) and numbers of epochs (200 to 500) tested, it can be attributed to the training data, which a large thyroid study could not separate from that of the method \citep{li2019diagnosis}.
\item \textbf{A comparison with radiologists and an expert-curated benchmark on the same nodules.} The automatically curated model lies within the AUC range of four radiologists and is not significantly different from a benchmark network trained on 1,278 expert-curated nodules (difference $-$0.036, 95\% CI: $-$0.083, 0.011), whereas the manually curated models fall significantly below that benchmark.
\item \textbf{Practical guidance.} Restricting training to the first stage of MADLaP, estimated to be more accurate (86\% vs 83\%), did not significantly change performance (pooled difference 0.015, 95\% CI: $-$0.006, 0.036), and the advantage held at every MADLaP batch size tested (256 to 1,024).
\item \textbf{A transparent analysis.} Every complete archived configuration is evaluated as a cross-validation ensemble, without selection on the test set, together with this study's own cross-validation, every training configuration and patient-clustered sensitivity analyses.
\end{itemize}

\section{Materials and Methods}
This retrospective study was approved by our institutional review board and was HIPAA compliant. A waiver of informed consent was also obtained.

\subsection{Overview}
The primary hypothesis of our study is the following: convolutional neural network-based classifiers for thyroid nodules in ultrasound, trained using large, automatically curated datasets perform better than the same classifiers trained using smaller, manually curated datasets.

In order to test this hypothesis, we created three training datasets from the development cohort (Manual Set, MADLaP Set, and S1 Set) following the workflow shown in Figure~\ref{fig:pipeline}. We trained the same CNN classifier on each dataset and evaluated model performance on an independent testing set. In addition to the primary comparison between the manual and automatically-curated datasets, we also compared the full MADLaP output to its Stage 1 subset (S1 Set), and explored the impact of batch size when training on automatically-curated data. The Manual Set represents the alternative use of the same expert annotations: the 378 patients whose images a radiologist curated are the patients on which MADLaP was developed \citep{zhang2023multistep}. Table~\ref{tab:demographics} summarizes the development cohort, the three training sets and the test set.

\subsection{Data}
\begin{itemize}
\item \textbf{Development cohort}: 3,981 patients who underwent thyroid FNA at our institution between October 2014 and October 2020.
\item \textbf{Manual Set}: radiologist-curated images of 378 randomly selected patients of the development cohort (803 images of 404 nodules).
\item \textbf{MADLaP Set}: the output of MADLaP run on all 3,981 patients of the development cohort (5,228 images of 2,605 nodules).
\item \textbf{S1 Set}: the part of the MADLaP Set produced by its first stage (4,150 images of 2,066 nodules).
\item \textbf{Test set}: 378 nodules of 320 patients from an earlier period, curated in a previous study \citep{weng2023deep}.
\end{itemize}

\begin{figure*}[t]
\centering
\includegraphics[width=\textwidth,alt={Study design diagram. Left: patient flows from the queried patients to the development cohort and to the test set, with the number excluded and the reason at each step and the biopsy period of each cohort. Middle: three lanes for the Manual, MADLaP and S1 training sets with how each was curated, the patients it came from, and its numbers of nodules, malignant nodules and images. Right: one shared box for the network, the cross-validation ensembles and the test-set evaluation.}]{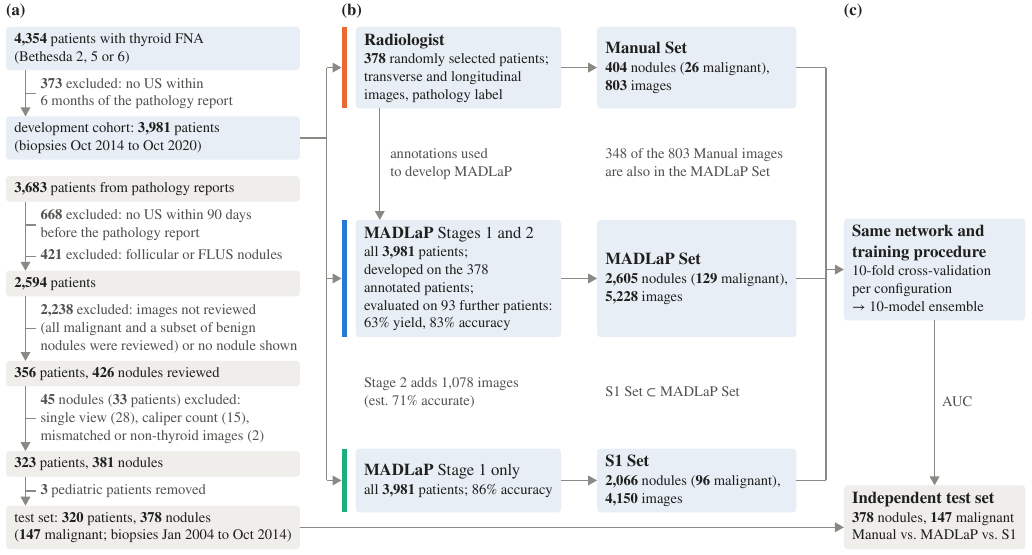}
\caption{Study design. (a) Development cohort \citep{zhang2023multistep} and test set \citep{weng2023deep}, with their biopsy periods and every exclusion with its reason. (b) The three training sets: the Manual Set (images curated by a radiologist for 378 randomly selected patients, on whom MADLaP was developed), the MADLaP Set (the output of MADLaP run on all 3,981 patients) and the S1 Set (the output of MADLaP's first stage). (c) The same network and procedure were used for every set; the ten fold models of each configuration were averaged into an ensemble and evaluated on the test set.}
\label{fig:pipeline}
\end{figure*}

\paragraph{Development cohort.} The study population included 4,354 patients from our health system. Subjects were identified using an internal search engine, querying pathology reports from thyroid fine needle aspirations (FNA) over a 6-year period (October 2014 to October 2020). Benign and malignant nodules were included based on the Bethesda system \cite{cibas2009bethesda}. Benign nodules were Bethesda 2, while malignant nodules were either Bethesda 5 (suspicious for malignancy) or 6 (diagnostic of malignancy). 373 patients were excluded due to the absence of ultrasound images within 6 months of the pathology report date. The remaining 3,981 patients formed the development cohort for this study.

The development cohort of 3,981 patients was used to create three training datasets. The overall design is shown in Figure~\ref{fig:pipeline}. To create a thyroid nodule dataset in the traditional way, patients who underwent thyroid US and US-guided FNA are identified, radiologists exclude some patients based on several criteria and select one transverse and one longitudinal image of each biopsied nodule, and the label of the images is determined from the FNA result in the pathology report \citep{buda2019management,zhang2023multistep}.

\paragraph{Manual Set.} The first dataset (referred to as the Manual Set) was based on a random selection of 402 patients from the development cohort, of whom 378 remained after 24 were excluded for no FNA at the included sites or no qualified ultrasound images \citep{zhang2023multistep}. These data were manually reviewed by a fellowship-trained radiologist (3 years post-fellowship experience at the time of annotation \citep{zhang2023multistep}). Transverse and longitudinal views were selected and manually labelled based on corresponding pathology reports. The total number of labelled images was 803 from 404 nodules (752 for benign nodules and 51 for malignant nodules), with both views for nearly every one of these nodules. The MADLaP study, which counted every nodule named in these patients' pathology reports, reported 462 nodules \citep{zhang2023multistep}.

\paragraph{MADLaP Set.} The second dataset (referred to as the MADLaP Set) was constructed using the automated tool from our prior work \citep{zhang2023multistep}: we employed a Multistep Automated Data Labeling Procedure (MADLaP) tool for automatic annotation. The tool employs rule-based NLP, optical character recognition (OCR), and a DL image segmentation model to analyze pathology reports, radiology reports, and data within US images to identify two target US images (transverse and longitudinal views) and assign corresponding pathology labels. The sequential (multistep) design of MADLaP allows us to evaluate each intermediate step. For this automatically-curated dataset, all 3,981 patients of the development cohort, including the 378 patients of the Manual Set, were run through MADLaP. This mirrors practice, where a curation tool is run on a whole archive, including any part that experts have already annotated; the two training sets therefore overlap, and 348 of the 803 Manual Set images also appear in the MADLaP Set. Just as the radiologist manually selected transverse and longitudinal images of a designated nodule in the Manual Set, MADLaP automatically selected two orthogonal images for nearly every nodule. The output from MADLaP consisted of 5,228 images from 2,605 nodules (4,970 images for benign nodules and 258 for malignant nodules). Both the manual and MADLaP image sets included images with calipers; in the former, the radiologist was instructed to select images with calipers, and in the latter, MADLaP used a pre-trained caliper detector \citep{zhang2023multistep,buda2019management} to find images with calipers.

\paragraph{MADLaP stages and label quality.} MADLaP is a multistep pipeline with two stages. In Stage 1, MADLaP searches focused thyroid FNA ultrasound studies (typically performed in the radiology department), which usually depict only the biopsied nodule and are more consistently labeled. If no such study is available, MADLaP proceeds to Stage 2, which searches the full diagnostic thyroid US examination and can further disambiguate candidate images using nodule measurements extracted from radiology reports \parencite{zhang2023multistep}.

MADLaP was developed on the 378 patients of the Manual Set and evaluated on a separate set of 93 patients (103 nodules) \citep{zhang2023multistep}.

We introduced yield rate and accuracy as our evaluation metrics to assess the efficiency and accuracy of MADLaP. Yield rate measures what percentage of the expected outcomes were identified by the model. Accuracy measures what percentage of the identified outcomes were matched with the ground-truths. Our experimental results have shown that MADLaP achieves 63\% of yield rate and 83\% accuracy on that set. Because an output was counted as incorrect whenever its image pair differed from the radiologist's selection, even when it showed the same nodule with the correct label, 17\% (11 of 65; exact 95\% CI of the error rate: 9\% to 28\%) is an upper bound on the label noise of the evaluated outputs \citep{zhang2023multistep}.

\paragraph{S1 Set.} The third dataset (referred to as the S1 Set) was created using only Stage 1 of the MADLaP pipeline. Since Stage 1 operates on more standardized imaging studies, this subset has a higher estimated label accuracy (86\%, 44 of 51, vs 83\%, 54 of 65, in the evaluation of MADLaP \citep{zhang2023multistep}) but a lower yield than the full MADLaP output. For the S1 Set, we have 4,150 images from 2,066 nodules in total. 3,958 images are for benign nodules, and 192 images are for malignant nodules. The S1 Set is a strict subset of the MADLaP Set, and the 1,078 images that Stage 2 adds had an estimated accuracy of 71\% (10 of 14 outputs in the evaluation of MADLaP \citep{zhang2023multistep}).

\paragraph{Test set.} For the testing part, 320 patients with 378 nodules were included from a test set published previously \citep{weng2023deep}. The testing set was collected from the electronic medical records of 3,683 patients identified through thyroid pathology reports at our institution between January 2004 and October 2014, a period that ends where that of the development cohort begins. Radiologists reviewed the images of all malignant nodules and of a subset of benign nodules, identifying 426 nodules in 356 patients; nodules with single-view, mismatched or non-thyroid images or with too many or too few calipers, and 3 pediatric patients, were then excluded \citep{weng2023deep}, which resulted in the final dataset of 320 patients with 378 nodules. For each nodule, one transverse image and one longitudinal image were selected. Benign nodules were Bethesda 2 (231 nodules) and malignant nodules Bethesda 5 or 6 (147 nodules, 38.9\%), and the nodules were imaged mainly with three scanner models (Philips iU22, 57.7\%; GE LOGIQ E9, 20.9\%; ATL HDI 5000, 16.9\%) \citep{weng2023deep}. An exhaustive comparison of image content, with every development image of the same frame size and with every training image as input to the network, found no test image among the training images (Appendix~\ref{app:sens}). All test results are reported on these 378 nodules with the published reference labels; the 383 nodules on which the models were originally evaluated give the same conclusions (Appendix~\ref{app:sens}).

\begin{table*}[t]
\centering
\small
\caption{Characteristics of the development cohort, the training sets and the test set.}
\label{tab:demographics}
\setlength{\tabcolsep}{4pt}
\begin{tabular}{@{}>{\raggedright\arraybackslash}p{3.0cm}>{\raggedright\arraybackslash}p{2.7cm}>{\raggedright\arraybackslash}p{2.4cm}>{\raggedright\arraybackslash}p{2.5cm}>{\raggedright\arraybackslash}p{2.5cm}>{\raggedright\arraybackslash}p{2.9cm}@{}}
\toprule
Characteristic & Development cohort & Manual Set & MADLaP Set & S1 Set & Test set \\
\midrule
Institution; biopsy period & Duke; \mbox{Oct 2014}--\mbox{Oct 2020} & Duke; development cohort & Duke; development cohort & Duke; development cohort & Duke; \mbox{Jan 2004}--\mbox{Oct 2014} \\
Patients & 3,981 (of 4,354 queried) & 378 (random) & run on all 3,981 & subset of MADLaP & 320 \\
Nodules & --- & 404 & 2,605 & 2,066 & 378 \\
Images (benign / malignant) & --- & 803 (752 / 51) & 5,228 (4,970 / 258) & 4,150 (3,958 / 192) & 756 (462 / 294) \\
Malignant nodules, \mbox{n (\%)} & --- & 26 (6.4) & 129 (5.0) & 96 (4.6) & 147 (38.9) \\
Image selection and label & --- & radiologist; pathology report & MADLaP Stages 1--2 & MADLaP Stage 1 & radiologists; cytopathology \\
Estimated label accuracy & --- & reference & 83\% (54 of 65)$^{*}$ & 86\% (44 of 51)$^{*}$ & reference \\
Reference standard & Bethesda 2 vs 5 or 6 (FNA cytology) & same & same & same & same \\
Scanners & --- & --- & --- & --- & iU22 57.7\%, \mbox{LOGIQ E9} 20.9\%, \mbox{HDI 5000} 16.9\%, other 4.5\% \\
Age, years, mean $\pm$ SD & not reported$^{\ddagger}$ & not reported$^{\dagger}$ & not reported$^{\dagger}$ & not reported$^{\dagger}$ & 52.7 $\pm$ 14.7 \\
Women & 3,505 of 4,354 queried (80.5\%)$^{\ddagger}$ & not reported$^{\dagger}$ & not reported$^{\dagger}$ & not reported$^{\dagger}$ & 304 of 378 nodules (80.4\%) \\
Maximum nodule size, cm, mean $\pm$ SD & --- & --- & --- & --- & 2.2 $\pm$ 1.2 \\
\bottomrule
\end{tabular}
\par\smallskip
\begin{minipage}{\textwidth}
\footnotesize
$^{*}$An output was counted as incorrect whenever its image pair differed from the radiologist's selection, so 17\% (14\% for Stage 1) is an upper bound on the label noise of the evaluated outputs \citep{zhang2023multistep}. $^{\dagger}$Demographics of the training sets were not available for this analysis. $^{\ddagger}$Version 1's sex counts (labelled there as the training dataset), which sum to the 4,354 patients found by the pathology-report search before exclusions; sex was not recorded for one. Ages are not reported because, in version 1's Table 1, the per-sex mean ages do not reproduce the overall mean. Test-set data from \citet{weng2023deep}; the models were originally evaluated on 383 nodules of 325 patients (Appendix~\ref{app:sens}).
\end{minipage}
\end{table*}

\subsection{Image Pre-processing}
Nodules were localized using a Faster Region-based Convolutional Neural Network as a detection algorithm \citep{ren2016faster}. We used the Faster Region-based Convolutional Neural Network with ResNet-101 backbone pre-trained on MS COCO dataset \citep{he2016deep,lin2014microsoft}. The algorithm was trained to detect measurement calipers included in images of thyroid nodules (standard clinical practice) \citep{buda2019management}. The calipers were then used to create bounding boxes that enclosed the entire nodule. After the detection process and a median filter, a square image of the corresponding nodule was extracted based on the location of the detected calipers with 32-pixel margin. The image was then padded to a square, resized into 160 $\times$ 160 pixels, contrast-stretched between its 1st and 99th intensity percentiles, denoised with fast non-local means and scaled to [$-$1, 1], following the public code of the benchmark study \citep{buda2019management}. The test images were the pre-processed images of the previous study \citep{weng2023deep}, prepared with the same steps.

\subsection{Deep Learning Algorithm for Nodule Classification}
\label{sec:network}
The network trunk, malignancy head, loss, optimizer, learning rate and augmentation were the same for all experiments, and every network was trained from scratch; batch size and number of epochs varied by configuration. The network is the multi-task convolutional neural network of the benchmark study \citep{buda2019management}, and its layers are listed in Appendix Table~\ref{tab:arch}. Its shared trunk has six convolutional layers with 3 $\times$ 3 kernels (8 to 48 filters) and ReLU activation functions. It also has five 2 $\times$ 2 max pooling layers and a 50\% dropout layer for regularization in the training phase. Finally, there are six output heads: a malignancy head with a sigmoid function and five auxiliary heads for ultrasound features (composition, echogenicity, shape, calcifications and margin). The output of the malignancy head is the probability of malignancy. Only the malignancy output was used in this study, and the score of a test nodule was the sum of the outputs for its transverse and longitudinal images (in cross-validation, the mean of its image outputs; Appendix~\ref{app:configs}). The targets of the five auxiliary heads were placeholders that did not correspond to our images (they had been copied by row position from the label table of the benchmark study), and they were handled in the same way for every training set. The network is compact (55,820 parameters), and its trunk, malignancy head and training procedure were the same for every training set; the auxiliary heads of the Manual Set models had slightly fewer output classes (Appendix Table~\ref{tab:arch}).

For training, we used a base learning rate of 0.001. The optimizer was Root Mean Squared Propagation. We implemented focal loss \citep{lin2017focal}, a dynamically scaled cross entropy loss, on every output head to tackle the class imbalance issue in the training dataset. The training data was augmented randomly with several operations, including flipping (horizontal and vertical), rotating and shearing (up to 15$^{\circ}$), translating (up to 20\%) and scaling (0.9 to 1.1). The archived models were trained with several different batch sizes and numbers of epochs: batch sizes of 128 and 256 for the Manual Set, 256, 512 and 1,024 for the MADLaP Set and 512 for the S1 Set, with 200 to 500 epochs (13 complete configurations; Appendix Table~\ref{tab:configs}). The number of epochs was fixed for each configuration; because several epoch variants had been compared on the test set in the original experiments, we report every archived configuration and select none of them. Each configuration was trained with 10-fold cross-validation on its training set, with folds split by nodule and stratified by malignancy, which gave ten fold models per configuration. For each configuration, we averaged the malignancy scores of its ten fold models to form a 10-model ensemble, which uses all training data and involves no choice made on the test set; this ensemble is the model we evaluate throughout, and we also report the mean and standard deviation of the ten individual fold models. All 13 configurations with ten archived fold models are reported; a fourteenth, with only three archived folds, is listed separately in Appendix Table~\ref{tab:configs}.

\subsection{Algorithm Comparison}
Deep learning models were trained based on the Manual Set, the MADLaP Set and the S1 Set. Once the models were trained (as shown in Figure~\ref{fig:pipeline}), we compared their performance. First, we tested whether the MADLaP-based algorithm (based on a large but automatically labelled dataset) could outperform the manual-based algorithm (based on a smaller but more carefully curated dataset). We compared the performance of the two models to determine if MADLaP is useful in improving the performance of deep learning algorithms. The primary comparison was the difference in test-set AUC between the MADLaP and Manual models trained for the same number of epochs (250; batch sizes of 512 and 256, with 2,500 and 750 parameter updates, respectively); as confirmation, we compared the pair of models trained with the same batch size (256; 300 epochs for MADLaP and 250 for Manual), the pooled ensembles of all models of the complete configurations of each set (90 MADLaP, 20 Manual and 20 S1 models), which involve no choice of configuration, and all 22 pairs of automatic (MADLaP or S1) and manual configurations. Because this comparison was designated in the present analysis, when the results of every configuration were already known, all of them are reported. The pair was designated before the training state stored with the models showed that its MADLaP model had used batch size 512; the pair with the same batch size gives a larger difference (Section~\ref{sec:primary}). Second, the MADLaP Set and the S1 Set algorithms were compared to evaluate whether a less accurate large dataset is more advantageous than a more accurate small dataset (a subset of the MADLaP dataset with a higher proportion of correct labels). Finally, we explored the impact of batch size on the automatically-curated dataset. A previous study suggested that increasing the batch size is a practical way of dealing with noisy labels \citep{rolnick2017deep}. Since the MADLaP dataset does not have perfect accuracy, we wanted to test whether increasing the batch size can significantly increase the performance of deep learning models trained on the automatically generated dataset. In addition, we compared the models with the four radiologists and the benchmark network that had been evaluated on the same test set in a previous study \citep{weng2023deep}.

\subsection{Statistical Analysis}
\label{sec:stats}
We calculated the performance of the deep learning algorithm based on the likelihood of malignancy outputted by the network. The metric we used is area under the receiver operating characteristic curve (AUC), computed at the nodule level. We calculated 95\% confidence intervals (CIs) with the method of DeLong et al. \citep{delong1988comparing}, and compared models evaluated on the same nodules with the paired DeLong test. Because some patients contributed more than one nodule, we repeated the comparisons between models (Tables~\ref{tab:main} and~\ref{tab:pairs}) and the comparisons of four models (the primary MADLaP model, both Manual models and the S1 model with the highest cross-validation AUC) with the radiologists, ACR TI-RADS and the benchmark (Table~\ref{tab:readers}) with patient-clustered variance estimates \citep{obuchowski1997nonparametric} and with a bootstrap that resampled patients (10,000 repetitions) \citep{efron1994introduction}; the single-fold-model comparison used the patient bootstrap only, and the comparisons of every configuration with the comparators (Section~\ref{sec:readers} and Table~\ref{tab:configs}) and the operating points (Table~\ref{tab:oppoints}) used nodule-level tests only. P values were Holm-adjusted \citep{holm1979simple} within each family: the 22 comparisons of automatic and manual configurations, and the two matched comparisons of the S1 and MADLaP Sets; a difference is called significant only when its adjusted P value is .05 or less, whereas statements that two models did not differ significantly use unadjusted P values and are given with CIs. Differences and their CIs were computed from unrounded AUCs, so a printed difference can differ by 0.001 from the difference of the printed AUCs. The radiologists were compared with the models through their five-point likelihood-of-malignancy scores, and ACR TI-RADS through the points assigned by three of them \citep{weng2023deep,wildmantobriner2024simplifying}; for each model, P values against these nine comparators, the benchmark included, were Holm-adjusted before any difference was called significant. For operating points, each model recommended biopsy for as many nodules as each set of radiologist recommendations, choosing its highest-scoring nodules, so that no outcome information was used to set a threshold; at equal numbers of biopsies, the difference in cancers detected equals the difference in benign biopsies, and these differences were tested with the exact McNemar test \citep{mcnemar1947note}, with Holm adjustment over the seven sets of recommendations of each model. From the DeLong standard error of the observed difference between the two primary models (0.015), a paired difference in AUC of 0.043 or more would be detected with 80\% power at a two-sided significance level of .05 (normal approximation; 0.045 with patient-clustered variance); for the MADLaP model against the benchmark or a radiologist, whose scores correlate less with the model's, the corresponding differences are 0.067 and 0.090 to 0.095. Two-sided P values less than or equal to .05, after Holm adjustment wherever a family of comparisons applied, were considered to indicate statistical significance. The networks were trained with TensorFlow/Keras 2.4, and statistical analysis used Python 3.12 with NumPy 2.5, SciPy 1.18, pandas 3.0 and scikit-learn 1.9. All archived fold models were re-run on a CPU with TensorFlow 2.16 for this analysis, and their predictions reproduced every per-fold test AUC recorded in the original experiments. A large language model was used to help write and run the analysis code for this version and to help refine the manuscript text; the authors reviewed the analyses and the text and take full responsibility for the content.

\section{Results}
\subsection{10-Fold Cross-Validation}
Evaluated by using 10-fold cross-validation, the deep learning algorithm achieved an AUC of 0.767 (95\% CI: 0.672, 0.863) for the manually selected dataset. For the MADLaP selected dataset, the deep learning algorithm achieved an AUC of 0.686 (95\% CI: 0.633, 0.738), also using 10-fold cross-validation. These are the pooled out-of-fold AUCs of the two configurations of the primary comparison; across all complete configurations, the values were 0.686 to 0.719 for the MADLaP Set, 0.742 to 0.749 for the S1 Set and 0.730 to 0.767 for the Manual Set (Appendix~\ref{app:configs}). Each value is measured against the set's own labels, so these estimates are not comparable between sets: label noise lowers the MADLaP estimate, the Manual estimate rests on only 26 malignant nodules, and the common test set provides the fair comparison. Averaged over each configuration's fold models, the image-level AUCs on their own training images, against each set's own labels, were 0.809 to 0.820 for the Manual Set, 0.733 to 0.759 for the MADLaP Set and 0.776 to 0.796 for the S1 Set, so the compact network fitted its training data only loosely, and the held-out AUC of the logged cross-validation run did not fall as training continued (Appendix Figure~\ref{fig:cv}).

\subsection{Manual Set vs.\ MADLaP Set}
\label{sec:primary}
\begin{figure}[t]
\centering
\includegraphics[width=\columnwidth,alt={ROC curves on the test set of the primary MADLaP and Manual ensembles and of the benchmark network, with the biopsy recommendations of four radiologists shown as filled points and the ACR TI-RADS recommendations of three of them shown as open points.}]{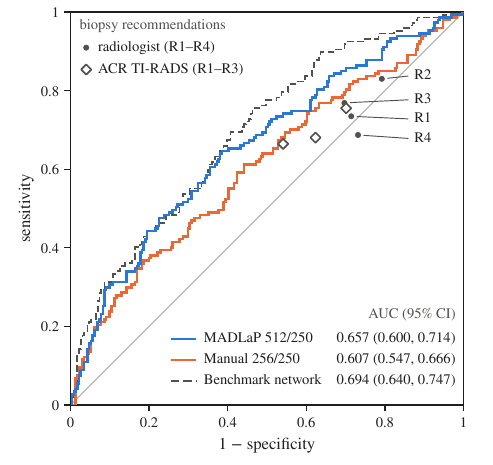}
\caption{ROC curves on the test set (378 nodules, 147 malignant; nodule level) of the primary MADLaP and Manual models (10-model ensembles; MADLaP 512/250, Manual 256/250) and of the benchmark network. Markers are the biopsy recommendations of four radiologists for the same nodules (filled) and the ACR TI-RADS recommendations of three of them (open; radiologist 3's on 377 nodules) \citep{weng2023deep,wildmantobriner2024simplifying}; they are decisions, not ROC curves. AUCs with 95\% DeLong CIs in the legend; the benchmark AUC was recomputed from its predictions (Table~\ref{tab:readers}).}
\label{fig:roc}
\end{figure}

For discriminating benign and malignant nodules in the testing set, with both datasets trained for 250 epochs (batch size 256 for the manually selected and 512 for the MADLaP selected dataset), the manually selected dataset has an AUC of 0.607 (95\% CI: 0.547, 0.666). It is significantly lower than the AUC of the MADLaP selected dataset, 0.657 (95\% CI: 0.600, 0.714; difference 0.051, 95\% CI: 0.020, 0.081; P~=~.001; Table~\ref{tab:main} and Figure~\ref{fig:roc}). With patient-clustered variance, the difference had a 95\% CI of 0.019 to 0.082 (P~=~.001), and a bootstrap that resampled patients gave 0.020 to 0.083. With the same batch size as the Manual model (256; 300 epochs), the MADLaP model had an AUC of 0.662 (difference 0.055, 95\% CI: 0.022, 0.088; P~=~.001). The pooled ensembles of all models of the complete configurations of each set (90 MADLaP, 20 Manual and 20 S1 models), which involve no choice of configuration, gave the same result (0.660 vs 0.611; difference 0.049, 95\% CI: 0.022, 0.077; P~<~.001). All 22 pairs of automatic and manual configurations favored automatic curation, with differences of 0.030 to 0.070, each with P~<~.05 (Figure~\ref{fig:robust}); after Holm adjustment, all 22 remained significant with DeLong variance (largest adjusted P~=~.048) and 20 with patient-clustered variance, the exceptions being the two pairs that involve the MADLaP configuration with batch size 512 and 340 epochs (adjusted P~=~.051); because the 22 pairs share two Manual ensembles and the same test nodules, this count shows consistency across configurations rather than independent replication. With a batch size of 512 and 350 epochs, the MADLaP model had an AUC of 0.666 (difference from the Manual model 0.059, 95\% CI: 0.026, 0.092; P~<~.001). The complete MADLaP configuration with the highest cross-validation AUC (batch size 1,024, 450 epochs), which was chosen without using the test set, had an AUC of 0.651, and its difference from the Manual configuration with the highest cross-validation AUC (batch size 256, 250 epochs) was 0.045 (95\% CI: 0.014, 0.076; P~=~.005). The advantage did not depend on ensembling: the ten fold models of the primary MADLaP and Manual configurations averaged 0.641 and 0.600 (difference 0.041, 95\% CI from a bootstrap that resampled patients: 0.013, 0.070), and 55 of the 90 MADLaP fold models exceeded the best of the 20 Manual fold models. These conclusions were unchanged on the 383 nodules evaluated originally and on the 381 nodules with radiologist scores, and the primary difference was unchanged at the image level (Appendix~\ref{app:sens}).

\begin{table*}[t]
\centering
\small
\caption{Test-set discrimination of the 10-model and pooled ensembles (378 nodules, 147 malignant).}
\label{tab:main}
\setlength{\tabcolsep}{4pt}%
\begin{tabular}{@{}llllll@{}}
\toprule
\begin{tabular}[b]{@{}l@{}}Model\\(batch/epochs)\end{tabular} & \begin{tabular}[b]{@{}l@{}}Training\\images\\(malignant)\end{tabular} & AUC (95\% CI) & \begin{tabular}[b]{@{}l@{}}Fold models,\\mean $\pm$ SD\end{tabular} & \begin{tabular}[b]{@{}l@{}}Difference vs\\Manual 256/250\\(95\% CI)\end{tabular} & P \\
\midrule
Manual 256/250 & 803 (51) & 0.607 (0.547, 0.666) & 0.600 $\pm$ 0.015 & \textcolor{gray}{\textit{reference}} & --- \\
\textbf{\begin{tabular}[t]{@{}l@{}}MADLaP 512/250\\(primary)\end{tabular}} & \textbf{5,228 (258)} & \textbf{0.657 (0.600, 0.714)} & \textbf{0.641 $\pm$ 0.016} & \textbf{0.051 (0.020, 0.081)} & \textbf{.001} \\
\begin{tabular}[t]{@{}l@{}}MADLaP 256/300\\(same batch size)\end{tabular} & 5,228 (258) & 0.662 (0.605, 0.719) & 0.638 $\pm$ 0.016 & 0.055 (0.022, 0.088) & .001 \\
MADLaP 512/350 & 5,228 (258) & 0.666 (0.608, 0.723) & 0.646 $\pm$ 0.026 & 0.059 (0.026, 0.092) & < .001 \\
\begin{tabular}[t]{@{}l@{}}MADLaP 1,024/450\\(highest CV AUC)\end{tabular} & 5,228 (258) & 0.651 (0.594, 0.708) & 0.635 $\pm$ 0.024 & 0.045 (0.014, 0.076) & .005 \\
\begin{tabular}[t]{@{}l@{}}S1 512/350\\(highest CV AUC)\end{tabular} & 4,150 (192) & 0.677 (0.621, 0.732) & 0.659 $\pm$ 0.014 & 0.070 (0.033, 0.107) & < .001 \\
\midrule
\begin{tabular}[t]{@{}l@{}}Pooled Manual\\(20 models)\end{tabular} & --- & 0.611 (0.551, 0.670) & --- & \textcolor{gray}{\textit{reference (pooled)}} & --- \\
\begin{tabular}[t]{@{}l@{}}Pooled MADLaP\\(90 models)\end{tabular} & --- & 0.660 (0.603, 0.717) & --- & 0.049 (0.022, 0.077) & < .001 \\
\begin{tabular}[t]{@{}l@{}}Pooled S1\\(20 models)\end{tabular} & --- & 0.675 (0.619, 0.731) & --- & 0.064 (0.031, 0.097) & < .001 \\
\bottomrule
\end{tabular}
\par\smallskip
\begin{minipage}{\textwidth}
\footnotesize
Paired DeLong tests with unadjusted P values (Holm-adjusted P of the configuration pairs: Table~\ref{tab:pairs}, at most .019 for the rows shown); the pooled rows are compared with the pooled Manual ensemble. Primary difference with patient-clustered variance: 95\% CI 0.019, 0.082, P~=~.001; patient-cluster bootstrap 0.020, 0.083. Pooled MADLaP excludes a configuration with three archived folds. Model labels give batch size/epochs; `same batch size': that of the Manual model. `Highest CV AUC': highest among the complete configurations, chosen by cross-validation alone, without the test set.
\end{minipage}
\end{table*}

\subsection{Comparison with Radiologists and the Benchmark}
\label{sec:readers}
On the same 378 nodules, the four radiologists had AUCs of 0.63 to 0.66 and the benchmark network, trained on 1,278 expert-curated nodules from another institution \citep{buda2019management,weng2023deep}, had an AUC of 0.69 \citep{weng2023deep}; the average of the four radiologists' scores, a composite that no individual reader matched, had an AUC of 0.718 (Table~\ref{tab:readers} and Figure~\ref{fig:teaser}). The MADLaP model was not significantly different from any of the radiologists (differences $-$0.004 to 0.030, with 95\% CIs between $-$0.070 and 0.094, each including 0; P~=~.35 to .91), from ACR TI-RADS points assigned by three of them (differences $-$0.025 to 0.032, with 95\% CIs between $-$0.094 and 0.110, each including 0; P~=~.41 to .51) or from the benchmark (difference $-$0.036, 95\% CI: $-$0.083, 0.011; P~=~.13). The benchmark is a single network \citep{buda2019management}, whereas each of our models is a 10-model ensemble; the single fold models of the primary MADLaP configuration averaged 0.641 (Table~\ref{tab:main}). Across all eleven automatic configurations, none differed significantly from any radiologist (differences $-$0.019 to 0.049, with 95\% CIs between $-$0.085 and 0.112, each including 0; all P~$\geq$~.12; Table~\ref{tab:configs}), only one (batch size 512, 340 epochs) had an unadjusted P value below .05 against the benchmark (P~=~.025), and two (batch size 512, 340 epochs, and batch size 1,024, 450 epochs) were lower than the averaged radiologist score with unadjusted P values below .05 (P~=~.019 and P~=~.036; Table~\ref{tab:configs}). The Manual models were also not significantly different from any individual radiologist (differences $-$0.055 to $-$0.015, with 95\% CIs between $-$0.121 and 0.052, each including 0; P~=~.11 to .65; Table~\ref{tab:configs}), but both fell significantly below the benchmark (P~=~.001) and below the averaged radiologist score (P~<~.001 and P~=~.001). The MADLaP model was numerically below the averaged radiologist score (difference $-$0.061, 95\% CI: $-$0.124, 0.002; P~=~.057).

In this all-biopsied test set, when each model recommended biopsy for as many nodules as the radiologists did, in each of their seven sets of recommendations (four by the radiologists' own judgment and three by ACR TI-RADS), the MADLaP model's recommendations included 7 to 18 more cancers than the radiologists' (and as many fewer benign nodules), and the Manual model's 1 to 14 more; none of these differences was significant after Holm adjustment (Appendix Table~\ref{tab:oppoints}).

\begin{table*}[t]
\centering
\small
\caption{Paired comparison with radiologists, ACR TI-RADS and the benchmark on the same 378 nodules.}
\label{tab:readers}
\setlength{\tabcolsep}{4pt}%
\begin{tabular}{@{}p{1.2in}lllll@{}}
\toprule
 &  & \multicolumn{2}{c}{\begin{tabular}[b]{@{}c@{}}MADLaP 512/250\\minus comparator\end{tabular}} & \multicolumn{2}{c}{\begin{tabular}[b]{@{}c@{}}Manual 256/250\\minus comparator\end{tabular}} \tabularnewline
\cmidrule(lr){3-4}\cmidrule(l){5-6}
Comparator & AUC (95\% CI) & \begin{tabular}[b]{@{}l@{}}Difference\\(95\% CI)\end{tabular} & P & \begin{tabular}[b]{@{}l@{}}Difference\\(95\% CI)\end{tabular} & P \tabularnewline
\midrule
\multicolumn{6}{@{}l}{\textit{Radiologists (5-point score)}} \tabularnewline
\raggedright \quad Radiologist 1 & 0.630 (0.585, 0.674) & 0.027 ($-$0.039, 0.094) & .42 & $-$0.023 ($-$0.093, 0.046) & .51 \tabularnewline
\raggedright \quad Radiologist 2 & 0.661 (0.609, 0.713) & $-$0.004 ($-$0.070, 0.062) & .91 & $-$0.055 ($-$0.121, 0.012) & .11 \tabularnewline
\raggedright \quad Radiologist 3 & 0.649 (0.601, 0.697) & 0.008 ($-$0.055, 0.071) & .80 & $-$0.042 ($-$0.109, 0.024) & .21 \tabularnewline
\raggedright \quad Radiologist 4 & 0.627 (0.582, 0.673) & 0.030 ($-$0.033, 0.093) & .35 & $-$0.021 ($-$0.088, 0.046) & .54 \tabularnewline
\addlinespace
\raggedright Average of the four radiologists' scores & 0.718 (0.666, 0.771) & $-$0.061 ($-$0.124, 0.002) & .057 & $-$0.112 ($-$0.177, $-$0.046) & < .001 \tabularnewline
\addlinespace
\multicolumn{6}{@{}l}{\textit{ACR TI-RADS points}} \tabularnewline
\raggedright \quad Radiologist 1 & 0.681 (0.627, 0.735) & $-$0.024 ($-$0.094, 0.047) & .51 & $-$0.074 ($-$0.146, $-$0.003) & .042 \tabularnewline
\raggedright \quad Radiologist 2 & 0.625 (0.568, 0.682) & 0.032 ($-$0.045, 0.110) & .41 & $-$0.018 ($-$0.097, 0.060) & .65 \tabularnewline
\raggedright \quad Radiologist 3 & 0.683 (0.629, 0.737) & $-$0.025 ($-$0.092, 0.043) & .48 & $-$0.076 ($-$0.147, $-$0.005) & .037 \tabularnewline
\addlinespace
\raggedright Benchmark network (1,278 expert-curated nodules) & 0.694$^{\dagger}$ (0.640, 0.747) & $-$0.036 ($-$0.083, 0.011) & .13 & $-$0.087 ($-$0.139, $-$0.035) & .001 \tabularnewline
\bottomrule
\end{tabular}
\par\smallskip
\begin{minipage}{\textwidth}
\footnotesize
Unadjusted paired DeLong P values. A difference is called significant only after Holm adjustment over the nine comparators of each model (Section~\ref{sec:stats}). Radiologist AUCs are empirical AUCs of five-point likelihood scores and reproduce the published binormal values to two decimals \citep{weng2023deep}; the mean of the four radiologists' AUCs is 0.642. ACR TI-RADS points were recomputed from three radiologists' recorded features \citep{wildmantobriner2024simplifying}; radiologist~3's points are missing for one nodule, so that row and its paired tests use the other 377 nodules. After Holm adjustment, only the Manual model's differences from the averaged score (P~=~.007) and from the benchmark (P~=~.009) are significant; every other adjusted P is .26 or more. Our models are 10-model ensembles; the benchmark is a single network. $^{\dagger}$Recomputed from the benchmark's predictions; published value 0.69 (0.64, 0.75) \citep{weng2023deep}.
\end{minipage}
\end{table*}

\subsection{MADLaP Set vs.\ S1 Set}
Comparing the two datasets in the same setup (same neural network architecture, 512 batch size and 350 epochs), the AUC of the MADLaP selected dataset is 0.666 (95\% CI: 0.608, 0.723). The AUC of MADLaP stage 1 dataset is 0.677 (95\% CI: 0.621, 0.732), which is not significantly different from the AUC of the total MADLaP selected dataset (difference 0.011, 95\% CI: $-$0.013, 0.036; P~=~.37). In the second matched configuration (340 epochs), the Stage 1 dataset was also higher (difference 0.025, 95\% CI: 0.0001, 0.050; P~=~.049), but neither matched comparison was significant after Holm adjustment across the two (adjusted P~=~.098 and .37), and the pooled ensembles did not differ significantly (difference 0.015, 95\% CI: $-$0.006, 0.036; P~=~.17). The Stage 1 dataset was therefore numerically higher in every comparison, but restricting training to it did not significantly change performance.

\subsection{Influence of Batch Size}
\begin{figure}[t]
\centering
\includegraphics[width=\columnwidth,alt={Forest plot of the difference in test-set AUC between each of the eleven automatic configurations and each of the two Manual configurations, with 95 percent confidence intervals, followed by two diamonds for the pooled ensembles of the MADLaP and S1 Sets, with a vertical line at zero.}]{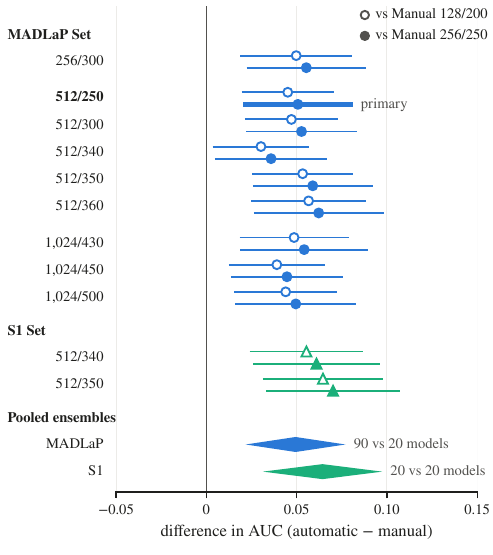}
\caption{Difference in test-set AUC (378 nodules) between every automatic configuration (MADLaP or S1; batch size/epochs) and each Manual configuration (open: 128/200; filled: 256/250), with 95\% DeLong CIs. Diamonds: the pooled MADLaP (90 models) and pooled S1 (20 models) ensembles of the complete configurations minus the pooled Manual ensemble (20 models). All 22 differences are positive and each has P~<~.05; after Holm adjustment, all 22 remain significant with DeLong variance and 20 with patient-clustered variance (Table~\ref{tab:pairs}). The 22 differences share two Manual ensembles and the same nodules, so they are correlated. Thick line: the primary comparison (MADLaP 512/250 minus Manual 256/250).}
\label{fig:robust}
\end{figure}

We trained multiple models with the three datasets using different batch sizes. The result is presented in Figure~\ref{fig:robust} and Appendix Figure~\ref{fig:folds}. The Manual Set models had AUCs of 0.612 and 0.607 at batch sizes of 128 and 256. The MADLaP Set models had an AUC of 0.662 at a batch size of 256, 0.642 to 0.669 at 512 and 0.651 to 0.661 at 1,024, and the pooled models trained with a batch size of 1,024 did not differ significantly from the model trained with 256 (difference $-$0.006, 95\% CI: $-$0.029, 0.018; P~=~.64). The S1 Set models, trained with a batch size of 512, had AUCs of 0.668 and 0.677. Every automatic configuration, at every batch size and number of epochs, outperformed both Manual models (Figure~\ref{fig:robust}). Models with the other batch sizes reported in version 1 of this preprint (below 256 for the MADLaP Set; below 512, and 1,024, for the S1 Set; and 32, 64 and 512 for the Manual Set) could not be recovered and are not analyzed (Appendix~\ref{app:changes}).

\section{Discussion}
The construction of a large thyroid nodule dataset is extremely time-consuming and costly \citep{willemink2020preparing,zhang2023multistep}. In addition, it requires the expertise of experienced radiologists and takes a very long time to select images from the database manually. Therefore, our study designed an experiment that shows the value of an automated dataset construction tool, MADLaP.

Automatically curated data produced a better classifier than the expert-curated data from which the curation tool was built. Among the available cases for training, only 378 out of 3,981 patients have annotations from radiologists. MADLaP was developed based on those 378 patients. 5,228 images were acquired from the 3,981 patients using MADLaP. The main objective of our experiment is to show that using the automatic dataset curating tool can achieve higher classification accuracy than using limited expert annotated data. In a real-life scenario, it is common that experienced radiologists annotate only a small portion of a thyroid nodule database. Our goal is to explore whether it is beneficial to develop an automatic tool using the annotated small dataset or one should simply use the annotated small dataset to train a classification network. We can see that the former has a significantly better performance by comparing the deep learning algorithm trained on the original 378 patients with the algorithm trained on the data acquired by MADLaP. Also, when we trained the two models with the same neural network structure and the same number of epochs (250), the model trained on the MADLaP generated dataset still has a significant advantage (AUC of 0.657 vs.\ 0.607, P~=~.001), and the advantage held for every pair of training configurations, including the pair trained with the same batch size of 256. In this way, we demonstrated the practicality of using automatically-curated data. The annotations of the same 378 patients yielded 6.5 times as many training images (5,228 vs 803) and 5.1 times as many malignant images (258 vs 51) when they were used to build a curation tool rather than a training set; evaluating the tool required annotating a further 93 patients \citep{zhang2023multistep}. This result fits a pattern seen across medical imaging, in which a large set of imperfect labels outperforms a small hand-labeled set \citep{fries2019weakly,dunnmon2020cross}, and labels extracted automatically from pathology reports trained a breast ultrasound system that reported a higher AUC than earlier systems built on small, manually labeled sets, although on different test sets \citep{shen2021artificial}.

We also compared our performance on the testing set to the benchmark. In the previous study, the benchmark performance on the testing set was 0.69 (95\% CI: 0.64, 0.75), and four radiologists achieved 0.63 to 0.66 \citep{weng2023deep}. Our MADLaP Set models reached 0.642 to 0.669; none was significantly different from any radiologist (Section~\ref{sec:readers}), and only one had an unadjusted P value below .05 against the benchmark (two against the averaged radiologist score), whereas the Manual Set models fell significantly below the benchmark. The benchmark deep learning algorithm was trained on 2,556 annotated thyroid nodule ultrasound images (1,278 nodules) from 1,139 patients at another institution \citep{buda2019management,weng2023deep}. Apart from the benchmark study's caliper detector, used to crop every model's images, our model only utilized annotated information from 378 patients, far less annotated data than the benchmark training dataset. We included 5,228 images, which were more images than the benchmark training dataset. However, our images were automatically-curated instead of manually annotated by expert radiologists. The benchmark was also trained with ACR TI-RADS feature labels for its auxiliary outputs \citep{buda2019management}, which our models did not have.

This is a difficult test set \citep{weng2023deep}: every nodule was biopsied, so its benign nodules were suspicious enough to be sampled, a case spectrum that can lower measured performance \citep{park2018methodologic,buda2019management}; it predates our training data by up to a decade; it includes images from a lower-resolution scanner model (ATL HDI 5000, 16.9\% of nodules) on which the benchmark reached an AUC of only 0.56 \citep{weng2023deep}, and about a fifth of its images come from frame sizes that never occur in our development data; and four radiologists reached AUCs of only 0.63 to 0.66 on it. The benchmark network itself fell from an AUC of 0.87 on its own test set to 0.69 on this one \citep{buda2019management,weng2023deep}, and about a quarter of externally validated deep learning algorithms for radiologic diagnosis lose 0.10 or more of their primary performance measure (AUC where reported) on external data \citep{yu2022external}. In the same health system, an FDA-cleared commercial system reached an AUC of 0.70 (95\% CI: 0.60, 0.81) on 649 nodules from one year of consecutive examinations, equal to that of the radiologists' clinical reports \citep{ashton2024assessment}; prospective cohorts have reported standalone AUCs of 0.75 and 0.82 \citep{li2023clinical,xu2024clinical}, and one Western external set reached 0.90 \citep{thomas2026agentic}. Most published AUCs come from cohorts in Asia: all 27 studies in a recent meta-analysis of externally validated models were conducted there \citep{gatta2026machine}, and in another meta-analysis the malignancy rate was a significant source of heterogeneity \citep{zhan2025diagnostic}.

To our knowledge, MADLaP is the only published pipeline that automatically produces nodule-specific, image-level, pathology-based labels for thyroid US, so it has no direct comparator; related thyroid work matches reports without selecting images \citep{lee2024leveraging} or learns from histology labels without image-level annotation \citep{hou2026pathological}. Nevertheless, it is meaningful to compare our experiments with other studies that also utilized noisy data since the automatically-curated dataset is not as accurate as a manually labeled dataset. One study done by Cheng Xue et al. showed that adding 5\% to 10\% of noise to the skin lesion dataset would only decrease the classification performance by 1.8 to 2.7 percentage points when the neural networks were trained with noise-robust methods, whereas standard training lost 3.7 to 7.2 points at the same noise levels and 11.1 points at 20\% noise \citep{xue2019robust}. Deep networks learn the consistent signal before they memorize mislabeled examples \citep{arpit2017closer}, and they tolerate many incorrect labels when enough correct examples are present \citep{rolnick2017deep}.

In our case, the MADLaP generated dataset has an estimated 83\% accuracy according to tests in the previous study \citep{zhang2023multistep}, which means we have at most around 17\% noisy data. In our case, we did not apply special techniques when training with the MADLaP generated dataset since we wanted to make sure the comparison between manually and automatically labeled datasets was fair. Therefore, we can safely conclude that MADLaP generated dataset, even with label noise estimated at up to 17\% and without any technique for handling it, is capable of training a better deep learning algorithm than the smaller manually selected dataset. At equal training-set size, label noise below about 10\% has shown no significant cost in binary medical image classification \citep{marini2025automatic}, and at 20\% or less in segmentation \citep{yu2020robustness}; in our setting, a 6.5-fold larger but noisier training set produced the better classifier. Because focal loss, like cross-entropy, is not robust to label noise \citep{ghosh2017robust,ma2020normalized}, and no technique for handling noise was used, the measured gain may underestimate what automatic curation can deliver with noise-robust training. Also, when we are training with only the Stage 1 yield of the MADLaP generated dataset, which was tested to have an 86\% accuracy in the previous study \citep{zhang2023multistep}, the performance is not significantly different from using the entire MADLaP generated dataset (pooled difference 0.015, 95\% CI: $-$0.006, 0.036), which is 26\% larger and has an estimated 3 percentage points more noise. Restricting the data to Stage 1 removes the 1,078 images added by Stage 2, of which an estimated 71\% were correctly matched \citep{zhang2023multistep}; the S1 Set was numerically higher in every comparison. Both outputs trained a better classifier than the expert-curated set, so, at MADLaP's accuracy, a strict accuracy filter was not needed for the curated data to be useful.

Our experiment also examined the impact of batch size on the automatically-curated medical dataset. On clean data, a larger batch size does not generally improve, and can impair, generalization \citep{keskar2017large,kandel2020effect}. As described in the study done by David Rolnick et al., increasing the batch size is a pragmatic method to reduce the negative impact of noisy labels \citep{rolnick2017deep}. They diluted each clean MNIST training example with 100 randomly labeled examples and still obtained a test accuracy above 90\%, and, in a separate experiment, larger batches made the network more robust to such noise. In our case, the thyroid nodule classification problem is much more difficult than MNIST \citep{deng2012mnist}. However, at the moderate noise level of the MADLaP dataset, we did not observe such an effect between batch sizes of 256 and 1,024: the MADLaP models had AUCs of 0.642 to 0.669 across these batch sizes, the pooled models at 1,024 and the model at 256 did not differ significantly (difference $-$0.006, 95\% CI: $-$0.029, 0.018), and the advantage over the manually curated dataset held at every MADLaP batch size. This is consistent with the much higher noise levels at which the effect was observed, where noisy labels outnumbered clean ones \citep{rolnick2017deep}, and it means that, within the range we tested, the benefit of automatic curation did not depend on tuning the batch size.

Automatic curation is also improving quickly: the best open language models now label radiology reports with F1 scores of 0.96 to 0.98 (0.99 in a rules-first hybrid) \citep{dorfner2024comparing,fink2026who}, and in thyroid care a language model matched nodules across ultrasound and pathology reports with an F1 score of 0.987 \citep{lee2024leveraging}. Because a larger automatically curated set produced a better classifier even at MADLaP's accuracy, we expect gains in curation yield, at equal or better accuracy, to translate into better models. Pretraining is a complementary route to label efficiency: networks pretrained on natural images have been fine-tuned for thyroid ultrasound \citep{chi2017thyroid}, and ultrasound foundation models pretrained without labels need fewer annotated images downstream \citep{jiao2024usfm}; neither supplies the labeled training data that automatic curation provides, and combining them requires retraining.

\section{Limitations}
Our study has several limitations. First, all training data came from one institution, and the test set came from the same institution, although from an earlier period, with a different scanner mix and with no test image found in any training set; patient-level overlap between the development and test cohorts could not be checked, and because the MADLaP Set draws on all 3,981 development patients whereas the Manual Set draws on 378 of them, any such overlap would more likely favor the MADLaP Set.
Second, batch sizes and numbers of epochs were explored in the original experiments with knowledge of test-set results; we therefore report every archived configuration, selection-free pooled ensembles and a configuration chosen by cross-validation alone, and several models described in version 1 (smaller batch sizes, the S1 Set at a batch size of 1,024 and the Manual Set at 512) could not be recovered. Third, cross-validation folds were split by nodule rather than by patient; the fold assignments behind the archived models were reconstructed from the original code and match, for every model, the number of validation images recorded with it. Fourth, the label accuracy of MADLaP was estimated on 93 patients, and that of the images added by Stage 2 on only 14 outputs; 17\% is an upper bound for the evaluated outputs, and the binary label noise of the training sets was not measured directly. Fifth, the auxiliary heads of the network were trained on placeholder targets in the same way for every training set; their effect on the comparison is unknown and may depend on the size of the training set. Sixth, the radiologists' scores come from an earlier reader study \citep{weng2023deep} and were not collected together with the models; the benchmark is a single network, whereas each of our models is a 10-model ensemble; the averaged score of the four readers was numerically higher than every model; five-point ratings produce ties that can lower empirical AUCs \citep{hanley1982meaning}, and when the MADLaP model's scores were coarsened to each radiologist's rating distribution, its AUC was 0.602 to 0.632, 0.025 to 0.030 below that of the matching radiologist, although not significantly (P~=~.33 to .40; Appendix~\ref{app:sens}); and with four readers we compared each reader individually rather than performing a multi-reader, multi-case analysis. Seventh, the malignancy prevalence differs between training (about 5\%; 6\% in the Manual Set) and testing (39\%), and every test nodule was biopsied; we therefore assessed discrimination only, did not assess calibration or clinical decision thresholds, and report operating points that describe this enriched cohort rather than clinical practice. Eighth, by design, the comparison of the Manual and MADLaP Sets changes data quantity and label quality together: it compares two uses of the same annotations and cannot separate the two factors. Ninth, the reference standard is cytology: Bethesda 5 nodules, counted as malignant, carry an average risk of malignancy of about 74\% \citep{ali2023bethesda}, and Bethesda 3 and 4 nodules were excluded from both the training and the test data. Tenth, the advantage of quantity over label quality is not universal: clean subsets have matched far larger noisy sets in face recognition \citep{wang2018devil}, and a few carefully annotated scans outperformed many noisily annotated ones in brain-lesion segmentation, although the reverse held for fetal-brain segmentation \citep{karimi2020deep}; in our setting, a binary task with moderate noise and a 6.5-fold size advantage, quantity prevailed. Finally, learning curves, noise-robust training, informative auxiliary targets, other network architectures, pretrained initialization and fine-tuning on the expert-curated set require retraining and are left to future work; because the compact network fitted its training data only loosely, the size of the advantage may differ for larger or pretrained networks.

\section{Conclusion}
Using the MADLaP pipeline to curate training data can significantly improve the performance of classification deep learning algorithms distinguishing malignant thyroid nodules. Both the full MADLaP output and its Stage-1 subset, estimated to be more accurate, trained better classifiers than the manually curated dataset, and the advantage held at every MADLaP batch size tested (256 to 1,024). This approach not only reduces the cost and time of manual annotation but also opens the door to scaling up deep learning development in medical imaging.

\section*{Data and code availability}
The patient images and labels cannot be shared. The MADLaP code is available at \url{https://github.com/MaciejMazurowski/MADLaP}, and the classification network follows the public code of the benchmark study (\url{https://github.com/MaciejMazurowski/thyroid-us}). Aggregate results and the analysis code are available from the corresponding author on request.

\section*{Acknowledgments}
We thank Jingxi Weng and the radiologists of the reader study of \citet{weng2023deep} for the test set and the reader scores.

{\small
\bibliographystyle{ieeenat_fullname}
\bibliography{references}
}

\appendix
\setcounter{figure}{0}
\setcounter{table}{0}
\renewcommand{\thefigure}{A\arabic{figure}}
\renewcommand{\thetable}{A\arabic{table}}
\renewcommand{\theHfigure}{A\arabic{figure}}
\renewcommand{\theHtable}{A\arabic{table}}

\section{Changes from version 1}
\label{app:changes}
This version replaces version 1 (v1; arXiv:2602.01020v1). Table~\ref{tab:changes} lists every substantive change. The main conclusion is unchanged; the conclusions about the S1 subset and the batch size are revised.

\begin{table*}[t]
\centering
\small
\caption{Changes from version 1.}
\label{tab:changes}
\setlength{\tabcolsep}{4pt}
\begin{tabular}{@{}>{\raggedright\arraybackslash}p{2.7cm}>{\raggedright\arraybackslash}p{4.4cm}>{\raggedright\arraybackslash}p{4.9cm}>{\raggedright\arraybackslash}p{4.4cm}@{}}
\toprule
Item & Version 1 & Version 2 & Basis \\
\midrule
Test-set AUC & For each configuration, the highest test AUC among its ten fold models (e.g., MADLaP 0.694, S1 0.689, the batch-size points) & 10-model ensemble for every configuration; fold-model mean $\pm$ SD; pooled ensembles & On the 383 nodules v1 used, the fold models whose test AUCs v1 reported exceeded their ensembles by 0.012 to 0.035 \\
Manual result at batch size 512 (0.643) & Reported & Not reported; the Manual models compared have batch sizes 128 and 256 & Model not archived \\
Confidence intervals and P values & About $\pm$0.03, bootstrap with unstated unit; ``P > .43'' & Nodule-level DeLong ($\pm$0.056 to 0.059 for the ensembles); paired DeLong; patient-clustered sensitivity analyses; Holm adjustment & v1 intervals could not be reproduced with a standard resampling unit; ``.43'' is close to a one-sided bootstrap share (0.41 to 0.46), not a two-sided P value (0.81 to 0.84) \\
Section 3.1 cross-validation & 0.78 (0.74, 0.82) for both sets, which is the cross-validation result of the benchmark study \citep{buda2019management} & This study's cross-validation for every configuration & --- \\
Epoch selection & ``selected based on ROCAUC in 10-fold cross validation'' & Epochs fixed per configuration; epoch variants had been compared on the test set, so all archived configurations are reported & Training records \\
S1 vs MADLaP & S1 slightly lower; ``use all the data'' & No significant difference; S1 numerically higher & Ensemble estimates \\
Batch size & AUC rose with batch size (0.656 to 0.700) & No significant difference between 256 and 1,024; other batch sizes not archived & v1 compared single best models, including runs that were not archived \\
Batch size of the MADLaP 512/250 model & Plotted at batch size 256 in the batch-size figure (0.673, one fold model); v1's own Manual vs MADLaP comparison used batch size 512 for both models (0.694 vs 0.643) & 512, from the training state; with Manual 256/250 (same number of epochs) it forms the primary comparison, and the MADLaP model with batch size 256 (300 epochs) is compared as a confirmation & Training state stored with each fold model (optimizer step count) \\
Test-set size & 378 nodules stated; computed on 383 & 378 nodules with published labels; 383 in Appendix~\ref{app:sens} & Consistency with \citet{weng2023deep} and the reader data \\
Network & Single sigmoid output & Six heads; auxiliary targets were placeholders; only the malignancy output evaluated & Training code \\
Manual Set & 802 images (50 malignant) & 803 images (51 malignant) from 404 nodules & Training files \\
\bottomrule
\end{tabular}
\end{table*}

\begin{table*}[t]
\ContinuedFloat
\centering
\small
\caption{Changes from version 1 (continued).}
\label{tab:changes-cont}
\setlength{\tabcolsep}{4pt}
\begin{tabular}{@{}>{\raggedright\arraybackslash}p{2.7cm}>{\raggedright\arraybackslash}p{4.4cm}>{\raggedright\arraybackslash}p{4.9cm}>{\raggedright\arraybackslash}p{4.4cm}@{}}
\toprule
Item & Version 1 & Version 2 & Basis \\
\midrule
Development period & 2013--2019 & October 2014 to October 2020 & \citet{zhang2023multistep} \\
Stage-1 accuracy & 88\% & 86\% (44 of 51) & \citet{zhang2023multistep} \\
Radiologist experience & 4 years post-fellowship & 3 years post-fellowship at the time of annotation & \citet{zhang2023multistep} \\
Benchmark AUC & 0.70 (0.64--0.75) & 0.69 (0.64, 0.75) & \citet{weng2023deep} \\
Annotated patients & 320 & 378 & 320 is the number of test-set patients \\
Table 1 & ``Training dataset''; age rows inconsistent & Characteristics of every set; sex of the 4,354 queried patients kept; ages not reported & Sex counts sum to the 4,354 queried patients; the per-sex mean ages do not reproduce the overall mean; the source table was not available \\
Figures & Screenshots with ultrasound thumbnails; network layer chain; batch-size chart & Vector figures without patient images; network in Table~\ref{tab:arch} & --- \\
References & Five misattributions; \citet{buda2019management} and \citet{weng2023deep} not cited & Corrected and extended & --- \\
Symbols & ``<'' and ``>'' printed as inverted punctuation & Corrected & Font encoding \\
Author order in the PDF & Yang, Zhang, Mazurowski, Wildman-Tobriner & Yang, Zhang, Wildman-Tobriner, Mazurowski, as in the arXiv listing & Authors \\
\bottomrule
\end{tabular}
\end{table*}

\section{All configurations and cross-validation}
\label{app:configs}
The archived models are the final-epoch models of each fold. We rebuilt the fold assignments with the original code, in which malignant nodules are assigned to folds in turn and benign nodules at random with a fixed seed, and scored each fold model on its held-out fold. In cross-validation, a nodule's score is the mean of its image outputs, because a few training nodules have one, three or four images.

For two configurations, the rebuilt held-out fold of one fold model reproduces its surviving training log exactly. For every fold model, the number of validation images recorded with the model equals the size of its rebuilt held-out fold under exactly one of the two seeds used in the original code, which identifies the fold assignment of every configuration (Table~\ref{tab:configs}).

A separate MADLaP Set run (batch size 128, 250 epochs), whose models were not archived, left learning curves for all ten folds, nine of them to 250 epochs and one to 210 epochs (Figure~\ref{fig:cv}a). Its held-out AUC rose from 0.589 after 10 epochs to a plateau of 0.668 to 0.679 from about 150 epochs (0.677 $\pm$ 0.064 at 250 epochs, nine folds).

A fourteenth configuration (MADLaP, batch size 1,024, 350 epochs) has only three archived fold models; it is listed but excluded from every comparison and pooled ensemble.

\begin{table*}[t]
\centering
\small
\caption{All trained configurations.}
\label{tab:configs}
\setlength{\tabcolsep}{4pt}%
\begin{tabular}{@{}llllllll@{}}
\toprule
 &  & \multicolumn{6}{c}{Test set (378 nodules)} \\
\cmidrule(l){3-8}
\begin{tabular}[b]{@{}l@{}}Set and\\batch/epochs\end{tabular} & \begin{tabular}[b]{@{}l@{}}CV pooled\\out-of-fold AUC\\(95\% CI)\end{tabular} & \begin{tabular}[b]{@{}l@{}}Ensemble AUC\\(95\% CI)\end{tabular} & \begin{tabular}[b]{@{}l@{}}Fold models,\\mean $\pm$ SD\end{tabular} & \begin{tabular}[b]{@{}l@{}}Highest\\fold\\model\end{tabular} & \begin{tabular}[b]{@{}l@{}}P vs\\radiologists\end{tabular} & \begin{tabular}[b]{@{}l@{}}P vs\\averaged\\score\end{tabular} & \begin{tabular}[b]{@{}l@{}}P vs\\benchmark\end{tabular} \\
\midrule
\multicolumn{8}{@{}l}{\textit{Manual Set}} \\
\quad 128/200 & 0.730 (0.633, 0.827) & 0.612 (0.553, 0.671) & 0.605 $\pm$ 0.020 & 0.632 & .15--.65 & .001 & .001 \\
\quad 256/250 & 0.767 (0.672, 0.863) & 0.607 (0.547, 0.666) & 0.600 $\pm$ 0.015 & 0.629 & .11--.54 & < .001 & .001 \\
\addlinespace
\multicolumn{8}{@{}l}{\textit{MADLaP Set}} \\
\quad 256/300 & 0.709 (0.660, 0.758)\textsuperscript{*} & 0.662 (0.605, 0.719) & 0.638 $\pm$ 0.016 & 0.660 & .28--.98 & .073 & .19 \\
\quad 512/250 & 0.686 (0.633, 0.738) & 0.657 (0.600, 0.714) & 0.641 $\pm$ 0.016 & 0.668 & .35--.91 & .057 & .13 \\
\quad 512/300 & 0.709 (0.660, 0.758)\textsuperscript{*} & 0.659 (0.602, 0.716) & 0.639 $\pm$ 0.026 & 0.685 & .32--.96 & .063 & .16 \\
\quad 512/340 & 0.712 (0.664, 0.760) & 0.642 (0.585, 0.700) & 0.627 $\pm$ 0.019 & 0.655 & .58--.84 & .019 & .025 \\
\quad 512/350 & 0.710 (0.659, 0.761) & 0.666 (0.608, 0.723) & 0.646 $\pm$ 0.026 & 0.690 & .23--.89 & .087 & .23 \\
\quad 512/360 & 0.692 (0.638, 0.746) & 0.669 (0.612, 0.725) & 0.646 $\pm$ 0.024 & 0.677 & .20--.82 & .13 & .31 \\
\quad 1,024/430 & 0.713 (0.664, 0.762) & 0.661 (0.604, 0.718) & 0.645 $\pm$ 0.024 & 0.698 & .30--.99 & .073 & .16 \\
\quad 1,024/450 & 0.719 (0.671, 0.767) & 0.651 (0.594, 0.708) & 0.635 $\pm$ 0.024 & 0.674 & .46--.94 & .036 & .099 \\
\quad 1,024/500 & 0.697 (0.649, 0.745) & 0.656 (0.599, 0.713) & 0.639 $\pm$ 0.025 & 0.674 & .37--.88 & .057 & .11 \\
\addlinespace
\multicolumn{8}{@{}l}{\textit{S1 Set}} \\
\quad 512/340 & 0.742 (0.688, 0.795) & 0.668 (0.612, 0.724) & 0.645 $\pm$ 0.027 & 0.692 & .22--.85 & .12 & .27 \\
\quad 512/350 & 0.749 (0.697, 0.801) & 0.677 (0.621, 0.732) & 0.659 $\pm$ 0.014 & 0.680 & .12--.64 & .19 & .45 \\
\midrule
\multicolumn{8}{@{}l}{\textit{Partial configuration (MADLaP Set)}} \\
\quad 1,024/350 & 0.721 (partial) & \multicolumn{6}{l}{\textit{3 archived fold models; not compared}} \\
\bottomrule
\end{tabular}
\par\smallskip
\begin{minipage}{\textwidth}
\footnotesize
Batch size/epochs agree with the training state stored with each fold model. Cross-validation (CV): pooled out-of-fold nodule-level AUC under the identified fold assignment of each configuration, scored against each set's own labels and therefore not comparable between sets; $^{*}$the rebuilt held-out fold of one fold model reproduces a surviving training log exactly. Test: 10-model ensembles on the 378 test nodules. P vs radiologists (range over the four), P vs averaged score and P vs benchmark: unadjusted paired DeLong tests.
\end{minipage}
\end{table*}

\begin{table*}[t]
\centering
\small
\caption{Automatic minus manual difference for all 22 configuration pairs.}
\label{tab:pairs}
\setlength{\tabcolsep}{3.5pt}%
\begin{tabular}{@{}lllllllll@{}}
\toprule
 & \multicolumn{4}{c}{vs Manual 128/200} & \multicolumn{4}{c}{vs Manual 256/250} \\
\cmidrule(lr){2-5}\cmidrule(l){6-9}
\begin{tabular}[b]{@{}l@{}}Automatic\\configuration\end{tabular} & \begin{tabular}[b]{@{}l@{}}Difference\\(95\% CI)\end{tabular} & P & \begin{tabular}[b]{@{}l@{}}Holm P,\\DeLong\end{tabular} & \begin{tabular}[b]{@{}l@{}}Holm P,\\clustered\end{tabular} & \begin{tabular}[b]{@{}l@{}}Difference\\(95\% CI)\end{tabular} & P & \begin{tabular}[b]{@{}l@{}}Holm P,\\DeLong\end{tabular} & \begin{tabular}[b]{@{}l@{}}Holm P,\\clustered\end{tabular} \\
\midrule
MADLaP 256/300 & \begin{tabular}[t]{@{}l@{}}0.050\\(0.019, 0.081)\end{tabular} & .002 & .015 & .013 & \begin{tabular}[t]{@{}l@{}}0.055\\(0.022, 0.088)\end{tabular} & .001 & .011 & .013 \\
\addlinespace[5pt]
\textbf{MADLaP 512/250} & \begin{tabular}[t]{@{}l@{}}0.045\\(0.020, 0.071)\end{tabular} & < .001 & .009 & .010 & \textbf{\begin{tabular}[t]{@{}l@{}}0.051\\(0.020, 0.081)\end{tabular}} & \textbf{.001} & \textbf{.011} & \textbf{.013} \\
MADLaP 512/300 & \begin{tabular}[t]{@{}l@{}}0.047\\(0.021, 0.073)\end{tabular} & < .001 & .006 & .009 & \begin{tabular}[t]{@{}l@{}}0.053\\(0.022, 0.083)\end{tabular} & < .001 & .009 & .013 \\
MADLaP 512/340 & \begin{tabular}[t]{@{}l@{}}0.030\\(0.003, 0.057)\end{tabular} & .027 & .048 & .051 & \begin{tabular}[t]{@{}l@{}}0.036\\(0.005, 0.067)\end{tabular} & .024 & .048 & .051 \\
MADLaP 512/350 & \begin{tabular}[t]{@{}l@{}}0.053\\(0.026, 0.081)\end{tabular} & < .001 & .004 & .004 & \begin{tabular}[t]{@{}l@{}}0.059\\(0.026, 0.092)\end{tabular} & < .001 & .009 & .010 \\
MADLaP 512/360 & \begin{tabular}[t]{@{}l@{}}0.057\\(0.025, 0.088)\end{tabular} & < .001 & .008 & .010 & \begin{tabular}[t]{@{}l@{}}0.062\\(0.026, 0.098)\end{tabular} & < .001 & .009 & .012 \\
\addlinespace[5pt]
MADLaP 1,024/430 & \begin{tabular}[t]{@{}l@{}}0.049\\(0.018, 0.079)\end{tabular} & .002 & .015 & .014 & \begin{tabular}[t]{@{}l@{}}0.054\\(0.019, 0.089)\end{tabular} & .003 & .018 & .020 \\
MADLaP 1,024/450 & \begin{tabular}[t]{@{}l@{}}0.039\\(0.012, 0.066)\end{tabular} & .004 & .019 & .021 & \begin{tabular}[t]{@{}l@{}}0.045\\(0.014, 0.076)\end{tabular} & .005 & .019 & .021 \\
MADLaP 1,024/500 & \begin{tabular}[t]{@{}l@{}}0.044\\(0.015, 0.073)\end{tabular} & .003 & .018 & .020 & \begin{tabular}[t]{@{}l@{}}0.050\\(0.016, 0.083)\end{tabular} & .004 & .019 & .021 \\
\midrule
S1 512/340 & \begin{tabular}[t]{@{}l@{}}0.055\\(0.024, 0.087)\end{tabular} & < .001 & .009 & .008 & \begin{tabular}[t]{@{}l@{}}0.061\\(0.026, 0.096)\end{tabular} & < .001 & .009 & .010 \\
S1 512/350 & \begin{tabular}[t]{@{}l@{}}0.065\\(0.031, 0.098)\end{tabular} & < .001 & .003 & .003 & \begin{tabular}[t]{@{}l@{}}0.070\\(0.033, 0.107)\end{tabular} & < .001 & .004 & .004 \\
\midrule
\multicolumn{9}{@{}l}{Positive differences: 22 of 22; unadjusted P $\le$ .05: 22 (DeLong), 22 (clustered).} \\
\multicolumn{9}{@{}l}{Holm-significant: 22 (DeLong), 20 (clustered).} \\
\bottomrule
\end{tabular}
\par\smallskip
\begin{minipage}{\textwidth}
\footnotesize
Differences in AUC on the 378 test nodules with 95\% DeLong CIs and unadjusted paired DeLong P values; Holm adjustment over the 22 pairs with DeLong variance and with patient-clustered variance \citep{obuchowski1997nonparametric}. MADLaP rows are grouped by batch size (batch size/epochs); bold: the primary comparison.
\end{minipage}
\end{table*}

\begin{figure*}[t]
\centering
\includegraphics[width=\textwidth,alt={Strip plot of the test AUC of the ten fold models, their mean and their ensemble with 95 percent confidence interval for each of the 13 complete configurations, grouped into the Manual, MADLaP and S1 Sets, with the range of the four radiologists as a grey band and the benchmark network as a dashed line.}]{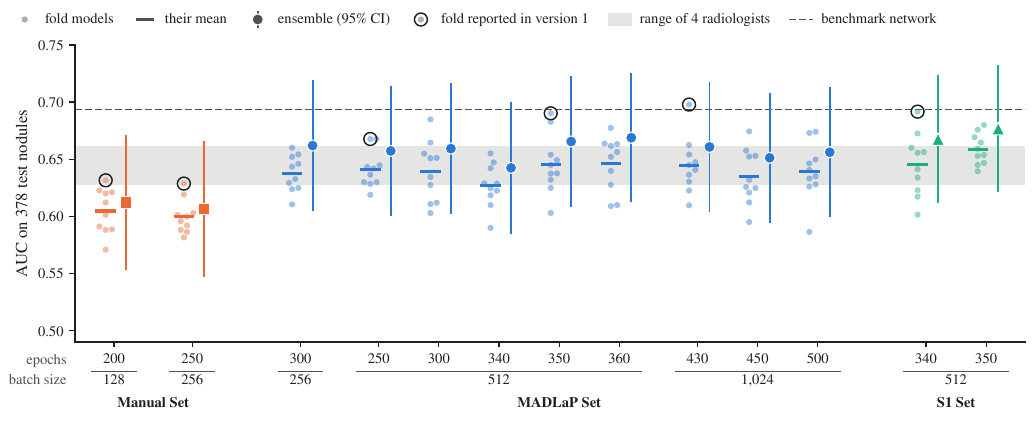}
\caption{Test AUC (378 nodules) of the ten fold models (dots), their mean (tick) and their ensemble (large marker, 95\% DeLong CI) for every complete configuration (batch size/epochs). Rings mark the fold model whose test AUC version 1 reported for that configuration; these fold models exceeded their ensembles by 0.010 to 0.037. Band: range of four radiologists; dashed line: benchmark network.}
\label{fig:folds}
\end{figure*}

\begin{figure*}[t]
\centering
\includegraphics[width=\textwidth,alt={Two panels. Left: held-out and training AUC by epoch in a 10-fold cross-validation run on the MADLaP Set, with one thin line per fold and a band for the mean and standard deviation. Right: pooled out-of-fold AUC with 95 percent confidence intervals for every complete configuration under its identified fold assignment.}]{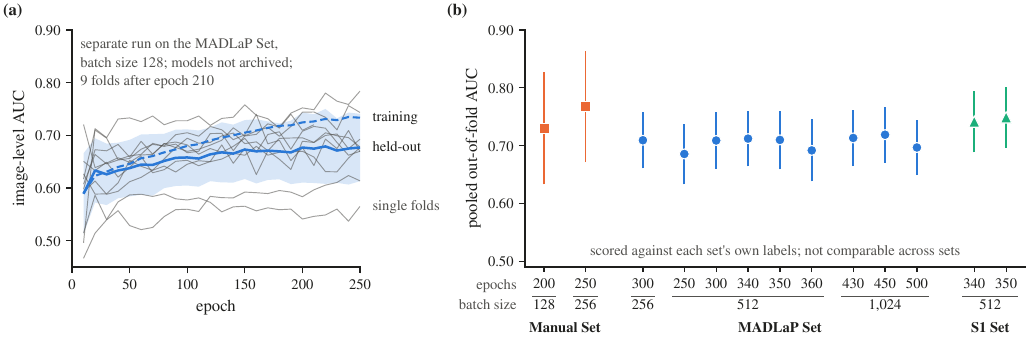}
\caption{Cross-validation. (a) Held-out image-level AUC during a 10-fold cross-validation run on the MADLaP Set (batch size 128, 250 epochs) whose models were not archived: folds (thin; one fold's log ends at 210 epochs), mean $\pm$ SD (band), training AUC (dashed). (b) Pooled out-of-fold nodule-level AUC (95\% DeLong CI) of every complete configuration under its identified fold assignment (404, 2,605 and 2,066 nodules for the Manual, MADLaP and S1 Sets). Each set is scored against its own labels.}
\label{fig:cv}
\end{figure*}

\section{Sensitivity analyses and operating points}
\label{app:sens}
All models were originally evaluated on 383 nodules (150 malignant) from 325 patients. These include the 378 nodules of the published test set; results on the 383 nodules and on the 381 nodules with radiologist scores lead to the same conclusions, and on the 377 nodules without label disagreements between the two sources, the primary and pooled results and the direction of all 22 pairs are unchanged, and 20 of the 22 pairs remain significant after Holm adjustment with DeLong variance (20 with patient-clustered variance; Table~\ref{tab:sens}).

Test images were searched for among the training images in two ways: the source frames located for 724 of the 756 test images were compared with every development frame of the same size, and every evaluated test crop (766 images) with every training image as input to the network (6,031 images). Neither search found a copy of a test image among the training images, whereas the same comparison of crops found all 348 Manual Set images that also appear in the MADLaP Set. Both searches detect copies that differ in intensity, compression or noise or are shifted by one pixel, but rarely copies cropped again with a shift of two or more pixels or rescaled by a few percent.

The primary difference is unchanged at the image level, and patient-clustered variance estimates \citep{obuchowski1997nonparametric} and a patient-cluster bootstrap give the same conclusions; with clustered variance, 20 of the 22 configuration pairs remain significant after Holm adjustment (Table~\ref{tab:pairs}).

The test-set nodule score is the sum of the outputs for the two views; their mean gives identical AUCs.

Choosing each set's configuration by cross-validation alone, without the test set, and comparing single fold models instead of ensembles both preserve the advantage (Table~\ref{tab:sens}).

Coarsening the model's scores to each radiologist's five-level rating distribution, which gives the model the same ties as the radiologist, lowers its AUC to 0.602 to 0.632, 0.025 to 0.030 below the matching radiologist, a difference that was not significant for any radiologist (P~=~.33 to .40; Table~\ref{tab:sens}).

Operating points (Table~\ref{tab:oppoints}): for each set of radiologist recommendations, each model recommended biopsy for the same number of nodules, its highest-scoring ones, without using outcome information. At equal numbers of biopsies, each additional cancer detected is one fewer benign biopsy.

\begin{table*}[t]
\centering
\small
\caption{Sensitivity analyses.}
\label{tab:sens}
\setlength{\tabcolsep}{4pt}%
\begin{tabular}{@{}p{1.55in}llp{2.55in}@{}}
\toprule
Analysis & Difference (95\% CI) & P & Note \tabularnewline
\midrule
\multicolumn{4}{@{}l}{\textit{MADLaP 512/250 minus Manual 256/250}} \tabularnewline
\raggedright \textcolor{gray}{\textit{Main analysis (378 nodules)}} & 0.051 (0.020, 0.081) & .001 & \raggedright \textcolor{gray}{\textit{paired DeLong}} \tabularnewline
\raggedright 383 nodules (original labels) & 0.053 (0.023, 0.084) & < .001 & \raggedright pooled MADLaP minus Manual 0.052 (0.024, 0.080), P~<~.001; positive pairs 22 of 22; Holm-significant 22 (DeLong), 22 (clustered) \tabularnewline
\raggedright 381 nodules with radiologist scores & 0.054 (0.024, 0.085) & < .001 & \raggedright pooled MADLaP minus Manual 0.054 (0.026, 0.082), P~<~.001; positive pairs 22 of 22; Holm-significant 22 (DeLong), 22 (clustered) \tabularnewline
\raggedright 377 nodules without label disagreement & 0.050 (0.019, 0.080) & .001 & \raggedright pooled MADLaP minus Manual 0.048 (0.020, 0.076), P~<~.001; positive pairs 22 of 22; Holm-significant 20 (DeLong), 20 (clustered) \tabularnewline
\raggedright Image level (756 images) & 0.049 (0.022, 0.076) & < .001 & \raggedright AUC 0.649 vs 0.600; nodule-cluster bootstrap CI; P from variance clustered by nodule \tabularnewline
\raggedright Patient-clustered variance & 0.051 (0.019, 0.082) & .001 & \raggedright Holm-significant pairs: 20 of 22 (largest adjusted P~=~.051) \tabularnewline
\raggedright Patient-cluster bootstrap & 0.051 (0.020, 0.083) & --- & \raggedright 10,000 resamples of patients; pair CIs excluding 0: 22 of 22 \tabularnewline
\raggedright Configurations chosen by cross-validation & 0.045 (0.014, 0.076) & .005 & \raggedright MADLaP 1,024/450 vs Manual 256/250, the highest cross-validation AUC among the complete configurations of each set \tabularnewline
\raggedright Single fold models & 0.041 (0.013, 0.070) & --- & \raggedright patient-cluster bootstrap CI; mean fold-model AUC 0.641 vs 0.600; 55 of 90 MADLaP fold models above the best Manual fold model (0.632) \tabularnewline
\midrule
\multicolumn{4}{@{}l}{\textit{Reader granularity: MADLaP 512/250 binned to each radiologist's rating frequencies, minus that radiologist}} \tabularnewline
\raggedright Radiologist 1 & $-$0.026 ($-$0.086, 0.034) & .40 & \raggedright binned model AUC 0.604; radiologist 0.630 \tabularnewline
\raggedright Radiologist 2 & $-$0.029 ($-$0.093, 0.035) & .38 & \raggedright binned model AUC 0.632; radiologist 0.661 \tabularnewline
\raggedright Radiologist 3 & $-$0.030 ($-$0.090, 0.030) & .33 & \raggedright binned model AUC 0.619; radiologist 0.649 \tabularnewline
\raggedright Radiologist 4 & $-$0.025 ($-$0.082, 0.032) & .39 & \raggedright binned model AUC 0.602; radiologist 0.627 \tabularnewline
\bottomrule
\end{tabular}
\par\smallskip
\begin{minipage}{\textwidth}
\footnotesize
Each row changes one element of the primary comparison (MADLaP 512/250 minus Manual 256/250); the reader-granularity rows compare the coarsened MADLaP model with each radiologist. P values are unadjusted.
\end{minipage}
\end{table*}

\begin{table*}[t]
\centering
\small
\caption{Operating points at the radiologists' numbers of biopsies.}
\label{tab:oppoints}
\setlength{\tabcolsep}{3pt}%
\begin{tabular}{@{}lllllllllll@{}}
\toprule
 &  &  & \multicolumn{4}{c}{MADLaP 512/250} & \multicolumn{4}{c}{Manual 256/250} \\
\cmidrule(lr){4-7}\cmidrule(l){8-11}
\begin{tabular}[b]{@{}l@{}}Radiologist\\decision set\end{tabular} & \begin{tabular}[b]{@{}l@{}}Biopsies n /\\nodules N\end{tabular} & \begin{tabular}[b]{@{}l@{}}Radiologist\\sensitivity /\\specificity\end{tabular} & \begin{tabular}[b]{@{}l@{}}Sensitivity /\\specificity\end{tabular} & \begin{tabular}[b]{@{}l@{}}Additional\\cancers\end{tabular} & \begin{tabular}[b]{@{}l@{}}McNemar\\P\end{tabular} & \begin{tabular}[b]{@{}l@{}}Holm\\P\end{tabular} & \begin{tabular}[b]{@{}l@{}}Sensitivity /\\specificity\end{tabular} & \begin{tabular}[b]{@{}l@{}}Additional\\cancers\end{tabular} & \begin{tabular}[b]{@{}l@{}}McNemar\\P\end{tabular} & \begin{tabular}[b]{@{}l@{}}Holm\\P\end{tabular} \\
\midrule
\multicolumn{11}{@{}l}{\textit{Radiologists' own recommendations}} \\
Radiologist 1 & 273 / 378 & 0.735 / 0.286 & 0.823 / 0.342 & 13 & .098 & .59 & 0.782 / 0.316 & 7 & .43 & > .99 \\
Radiologist 2 & 305 / 378 & 0.830 / 0.208 & 0.878 / 0.238 & 7 & .31 & > .99 & 0.844 / 0.216 & 2 & .87 & > .99 \\
Radiologist 3 & 274 / 378 & 0.769 / 0.303 & 0.830 / 0.342 & 9 & .23 & > .99 & 0.782 / 0.312 & 2 & .89 & > .99 \\
Radiologist 4 & 270 / 378 & 0.687 / 0.268 & 0.810 / 0.346 & 18 & .027 & .19 & 0.782 / 0.329 & 14 & .10 & .72 \\
\addlinespace
\multicolumn{11}{@{}l}{\textit{ACR TI-RADS recommendations}} \\
Radiologist 1 & 244 / 378 & 0.680 / 0.377 & 0.748 / 0.420 & 10 & .25 & > .99 & 0.721 / 0.403 & 6 & .53 & > .99 \\
Radiologist 2 & 273 / 378 & 0.755 / 0.299 & 0.823 / 0.342 & 10 & .21 & > .99 & 0.782 / 0.316 & 4 & .68 & > .99 \\
Radiologist 3 & 222 / 377 & 0.664 / 0.459 & 0.719 / 0.494 & 8 & .35 & > .99 & 0.671 / 0.463 & 1 & > .99 & > .99 \\
\bottomrule
\end{tabular}
\par\smallskip
\begin{minipage}{\textwidth}
\footnotesize
Thresholds set without outcome information (Section~\ref{sec:stats}): each model recommended biopsy for the n highest-scoring of the N nodules. Additional cancers: cancers detected by the model minus those detected by the radiologist; at equal numbers of biopsies, each additional cancer detected is one fewer benign biopsy. Results describe this all-biopsied, 39\%-malignant test set and are not estimates for clinical practice. McNemar P: exact test, unadjusted; Holm P: adjusted over the seven decision sets of each model.
\end{minipage}
\end{table*}

\section{Network}
\label{app:arch}
Table~\ref{tab:arch} lists the layers of the network \citep{buda2019management}.

\begin{table*}[t]
\centering
\footnotesize
\caption{Network.}
\label{tab:arch}
\setlength{\tabcolsep}{3pt}
\begin{tabular}{@{}l l l r@{}}
\toprule
Block & Operation & Output & Parameters \\
\midrule
Input & --- & 160$\times$160$\times$1 & --- \\
Conv 1 & 3$\times$3, 8; pool & 80$\times$80$\times$8 & 80 \\
Conv 2 & 3$\times$3, 12; pool & 40$\times$40$\times$12 & 876 \\
Conv 3 & 3$\times$3, 16; pool & 20$\times$20$\times$16 & 1,744 \\
Conv 4 & 3$\times$3, 24; pool & 10$\times$10$\times$24 & 3,480 \\
Conv 5 & 3$\times$3, 32; pool & 5$\times$5$\times$32 & 6,944 \\
Conv 6 & 3$\times$3, 48 & 5$\times$5$\times$48 & 13,872 \\
Dropout & rate 0.5 & 5$\times$5$\times$48 & 0 \\
Trunk total & & & 26,996 \\
\midrule
Malignancy & 5$\times$5, sigmoid & 1 & 1,201 \\
Composition & 5$\times$5 & 6 & 7,206 \\
Echogenicity & 5$\times$5 & 6 & 7,206 \\
Shape & 5$\times$5 & 1 & 1,201 \\
Calcifications & 5$\times$5 & 5 & 6,005 \\
Margin & 5$\times$5 & 5 & 6,005 \\
\midrule
Total & & & 55,820 \\
\bottomrule
\end{tabular}
\par\smallskip
\begin{minipage}{\columnwidth}
\footnotesize
Trunk: 3$\times$3 convolutions with the given number of filters, each followed by ReLU; pool: 2$\times$2 max pooling. Heads: 5$\times$5 convolutions; output: number of classes. Manual Set models have 5-, 5- and 4-class composition, echogenicity and margin heads (52,217 parameters). Auxiliary-head targets were placeholders (Section~\ref{sec:network}); only the malignancy output was evaluated. Loss: focal loss ($\gamma$ = 2; $\alpha$ = 2, a constant factor applied to both classes) on every head, weights 1.0; RMSprop, learning rate 0.001.
\end{minipage}
\end{table*}

\end{document}